# Cross-lingual Transfer of Sentiment Classifiers

Marko Robnik-Šikonja[1], Kristjan Reba[1], Igor Mozetič[2]

[1]University of Ljubljana, Faculty of Computer and Information Science
Večna pot 113, SI-1000 Ljubljana, Slovenia
[marko.robnik@fri.uni-lj.si](marko.robnik@fri.uni-lj.si)     [kr3377@student.uni-lj.si](kr3377@student.uni-lj.si)

[2]Jožef Stefan Institute
Jamova 39, SI-1000 Ljubljana, Slovenia
[igor.mozetic@ijs.si](igor.mozetic@ijs.si)

**Abstract**

Word embeddings represent words in a numeric space so that semantic relations between words are represented as distances and directions in the vector space. Cross-lingual word embeddings transform vector spaces of different languages so that similar words are aligned. This is done by constructing a mapping between vector spaces of two languages or learning a joint vector space for multiple languages. Cross-lingual embeddings can be used to transfer machine learning models between languages, thereby compensating for insufficient data in less-resourced languages. We use cross-lingual word embeddings to transfer machine learning prediction models for Twitter sentiment between 13 languages. We focus on two transfer mechanisms that recently show superior transfer performance. The first mechanism uses the trained models whose input is the joint numerical space for many languages as implemented in the LASER library. The second mechanism uses large pretrained multilingual BERT language models. Our experiments show that the transfer of models between similar languages is sensible, even with no target language data. The performance of cross-lingual models obtained with the multilingual BERT and LASER library is comparable, and the differences are language-dependent. The transfer with CroSloEngual BERT, pretrained on only three languages, is superior on these and some closely related languages.

1. **INTRODUCTION**

Word embeddings are representations of words in numerical form, as vectors of typically several hundred dimensions. The vectors are used as inputs to machine learning models; for complex language processing tasks, these generally are deep neural networks. The embedding vectors are obtained from specialised neural network-based embedding algorithms, e.g., fastText (Bojanowski et al., 2017) for morphologically-rich languages. Word embedding spaces exhibit similar structures across languages, even when considering distant language pairs like English and Vietnamese (Mikolov et al., 2013). This means that embeddings independently produced from monolingual text resources can be aligned, resulting in a common cross-lingual representation, called cross-lingual embeddings, which allows for fast and effective integration of information in different languages.

There exist several approaches to cross-lingual embeddings. The first group of approaches uses monolingual embeddings with optional help from a bilingual dictionary to align the pairs of embeddings (Artetxe et al., 2018a). The second group of approaches uses bilingually aligned (comparable or even parallel) corpora to construct joint embeddings (Artetxe and Schwenk, 2019). This approach is implemented in the LASER library[1] and is available for 93 languages. The third type of approaches is based on large pretrained multilingual masked language models such as BERT (Devlin et al., 2019). In this work, we focus on the second and third group of approaches. In particular, from the third group, we apply two variants of BERT models, the original multilingual BERT model (mBERT), trained on 104 languages, and trilingual CroSloEngual BERT (Ulčar and Robnik-Šikonja, 2020) trained on Croatian, Slovene, and English (CSE BERT).

Sentiment annotation is a costly and lengthy operation, with a relatively low inter-annotator agreement (Mozetič et al., 2016). Large annotated sentiment datasets are, therefore, rare, especially for low-resourced languages. The transfer of already trained models or datasets from other languages would increase the ability to study sentiment-related phenomena for many more languages than possible today.

Our study aims to analyse the abilities of modern cross-lingual approaches for the transfer of trained models between languages. We study two cross-lingual transfer technologies, using a joint vector space computed from parallel corpora with the LASER library and multilingual BERT models. The advantage of our study is sizeable comparable classification datasets in 13 different languages, which gives credibility and general validity to our findings. Further, due to the datasets' size, we can reliably test different transfer modes: direct transfer between languages (called a zero-shot transfer) and transfer with enough fine-tuning data in the target language. In the experiments, we study two cross-lingual transfer modes based on projections of sentences into a joint vector space. The first mode transfers trained models from source to target languages. A model is trained on the source language(s) and used for classification in the target language(s). This model transfer is possible because texts in all processed languages are embedded into the common vector space. The second mode expands the training set with instances from other languages, and then all instances are mapped into the common vector space during neural network training. Besides the cross-lingual transfer, we analyse the quality of representations for the Twitter sentiment classification and compare the common vector space for several languages constructed by the LASER library, multilingual BERT models, and the traditional bag-of-words approach. The results show a relatively low decrease in predictive performance when transferring trained sentiment prediction models between similar languages and superior performance of multilingual BERT models covering only three languages.

---

[1] https://github.com/facebookresearch/LASER

The paper is divided into four more sections. In Section 2, we present background on different types of cross-lingual embeddings: alignment of monolingual embeddings, building a common explicit vector space for several languages, and large pretrained multilingual contextual models. We also discuss related work on Twitter sentiment analysis and cross-lingual transfer of classification models. In Section 3, we present a large collection of tweets from 13 languages used in our empirical evaluation, the implementation details of our deep neural network prediction models, and the evaluation metrics used. Section 4 contains four series of experiments. We first evaluate different representation spaces and compare the LASER common vector space with multilingual BERT models and conventional bag-of-ngrams. We then analyse the transfer of trained models between languages from the same language group and from a different language group, followed by expanding datasets with instances from other languages. In Section 5, we summarise the results and present ideas for further work.

## 2. BACKGROUND AND RELATED WORK

Word embeddings represent each word in a language as a vector in a high dimensional vector space so that the relations between words in a language are reflected in their corresponding embeddings. Cross-lingual embeddings attempt to map words represented as vectors from one vector space to another so that the vectors representing words with the same meaning in both languages are as close as possible. Søgaard et al. (2019) present a detailed overview and classification of cross-lingual methods.

Cross-lingual approaches can be sorted into three groups, described in the following three subsections. The first group of methods uses monolingual embeddings with (an optional) help from bilingual dictionaries to align the embeddings. The second group of approaches uses bilingually aligned (comparable or even parallel) corpora for joint construction of embeddings in all handled languages. The third type of approaches is based on large pretrained multilingual masked language models such as BERT (Devlin et al., 2019). In contrast to the first two types of approaches, the multilingual BERT models are typically used as starting models, which are fine-tuned for a particular task without explicitly extracting embedding vectors.

In Section 2.1, we first present background information on the alignment of individual monolingual embeddings. We describe the projections of many languages into a joint vector space in Section 2.2, and in Section 2.3, we present variants of multilingual BERT models. In Section 2.4, we describe related work on Twitter sentiment classification. Finally, in Section 2.5, we outline the related work on the cross-lingual transfer of classification models.

### 2.1. Alignment of monolingual embeddings

Cross-lingual alignment methods take precomputed word embeddings for each language and align them with the optional use of bilingual dictionaries. Two types of monolingual embedding alignment methods exist. The first type of approaches map vectors representing words in one of the languages into the vector space of the other language (and vice-versa). The second type of approaches maps embeddings from both languages into a joint vector space. The goal of both types of alignments is the same: the embeddings for words with the same meaning must be as close as possible in the final vector space. A comprehensive summary of existing approaches can be found in (Artetxe et al., 2018a). The open-source *vecmap*[2] library contains implementations of methods described in (Artetxe et al., 2018a) and can align monolingual embeddings using a supervised, semi-supervised, or unsupervised approach.

---
[2] https://github.com/artetxem/vecmap

The supervised approach requires a bilingual dictionary, which is used to match embeddings of equivalent words. The embeddings are aligned using the Moore-Penrose pseudo-inverse, which minimises the sum of squared Euclidean distances. The algorithm always converges but can be caught in a local maximum. Several methods (e.g., stochastic dictionary introduction or frequency-based vocabulary cut-off) are used to help the algorithm climb out of local maxima. A more detailed description of the algorithm is given in (Artetxe et al., 2018b).

The semi-supervised approach uses a small initial seeding dictionary, while the unsupervised approach is run without any bilingual information. The latter uses similarity matrices of both embeddings to build an initial dictionary. This initial dictionary is usually of low but sufficient quality for later processing. After the initial dictionary (either by seeding dictionary or using similarity matrices) is built, an iterative algorithm is applied. The algorithm first computes optimal mapping using the pseudo-inverse approach for the given initial dictionary. The optimal dictionary for the given embeddings is then computed, and the procedure iterates with the new dictionary.

When constructing mappings between embedding spaces, a bilingual dictionary can help as its entries are used as anchors for the alignment map for supervised and semi-supervised approaches. However, lately, researchers have proposed methods that do not require a bilingual dictionary but rely on the adversarial approach (Conneau et al., 2018) or use the words' frequencies (Artetxe et al., 2018b) to find a required transformation. These are called unsupervised approaches.

### 2.2. Projecting into a joint vector space

To construct a common vector space for all the processed languages, one requires a large aligned bilingual or multilingual parallel corpus. The constructed embeddings must map the same words in different languages as close as possible in the common vector space. The availability and quality of alignments in the training set corpus may present an obstacle. While Wikipedia, subtitles, and translation memories are good sources of aligned texts for large languages, less-resourced languages are not well-presented and building embeddings for such languages is a challenge.

LASER (Language-Agnostic SEntence Representations) is a Facebook research project focusing on joint sentence representation for many languages (Artetxe and Schwenk, 2019). Strictly speaking, LASER is not a word but a sentence embedding method. Similar to machine translation architectures, LASER uses an encoder-decoder architecture. The encoder is trained on a large parallel corpus, translating a sentence in any language or script to a parallel sentence in either English or Spanish (whichever exists in the parallel corpus), thereby forming a joint representation of entire sentences in many languages in a shared vector space. The project focused on scaling to many languages; currently, the encoder supports 93 different languages. Using LASER, one can train a classifier on data from just one language and use it on any language supported by LASER. A vector representation in the joint embedding space can be transformed back into a sentence using a decoder for the specific language.

### 2.3. Multilingual BERT and CroSloEngual BERT

BERT (Bidirectional Encoder Representations from Transformers) embedding (Devlin et al., 2019) generalises the idea of a language model (LM) to masked LMs, inspired by the cloze test, which checks understanding of a text by removing a few words, which the participant is asked to replace. The masked LM randomly masks some of the tokens from the input, and the task is to predict the missing token based on its neighbourhood. BERT uses transformer neural networks (Vaswani et al., 2017) in a bidirectional sense and further introduces the task of predicting whether two sentences appear in a sequence. The input representation of BERT is a sequence of tokens representing sub-word units. The input is constructed by summing the embeddings of corresponding tokens, segments, and positions. Some widespread words are kept as single tokens; others are split into sub-words (e.g., frequent stems, prefixes, suffixes—if needed down to single letter tokens). The original BERT project offers pre-trained English, Chinese, and multilingual model. The latter, called mBERT, is trained on 104 languages simultaneously.

To use BERT in classification tasks only requires adding connections between its last hidden layer and new neurons corresponding to the number of classes in the intended task. The fine-tuning process is applied to the whole network, and all the parameters of BERT and new class-specific weights are fine-tuned jointly to maximise the log-probability of correct labels.

Recently, a new type of multilingual BERT models emerged that reduce the number of languages in multilingual models. For example, CSE BERT (Ulčar & Robnik-Šikonja, 2020) uses Croatian, Slovene (two similar less-resourced languages from the same language family), and English. The main reasons for this choice are to represent each language better and keep sensible sub-word vocabulary, as shown by Virtanen et al. (2019). This model is built with the cross-lingual transfer of prediction models in mind. As CSE BERT includes English, we expect that it will enable a better transfer of existing prediction models from English to Croatian and Slovene.

### 2.4. Twitter sentiment classification

We present a brief overview of the related work on automated sentiment classification of Twitter posts. We summarise the published labelled sets used for training the classification models and the machine learning methods applied for training. Most of the related work is limited to only English texts.

To train a sentiment classifier, one needs a reasonably large training dataset of tweets already labelled with the sentiment. One can rely on a proxy, e.g., emoticons used in the tweets, to determine the intended sentiment; however, high-quality labelling requires the engagement of human annotators. There exist several publicly available and manually labelled Twitter datasets. They vary in the number of examples from several hundred to several thousand, but to the best of our knowledge, so far, none exceeds 20,000 entries. Saif et al. (2013) describe eight Twitter sentiment datasets and introduce a new one that contains separate sentiment labels for tweets and entities. Rosenthal et al. (2015) provide statistics for several of the 2013–2015 SemEval datasets.

There are several supervised machine learning algorithms suitable to train sentiment classifiers from sentiment-labelled tweets. For example, in the SemEval-2015 competition, before the rise of deep neural networks, the most often used algorithms for the sentiment analysis on Twitter (Rosenthal et al., 2015) were support vector machines (SVM), maximum entropy, conditional random fields, and linear regression. In other cases, frequently used classifiers were naive Bayes, k-nearest neighbours, and even decision trees. Often, SVM was shown as the best performing classifier for the Twitter sentiment. However, only recently, when researchers started to apply deep learning for the Twitter sentiment classification, considerable improvements in classification performance were observed (Wehrmann et al., 2017; Jianqiang et al., 2018; Naseem et al., 2020). Similarly to our approach, recent approaches use contextual embeddings such as ELMo (Peters et al., 2018) and BERT (Devlin et al., 2019), but in a monolingual setting.

## 2.5. Transfer of trained models

Cross-lingual word embeddings can be used directly as inputs in natural language processing models. The main idea is to train a model on data from one language and then apply it to another, relying on shared cross-lingual representation. Several tasks have been attempted in the testing cross-lingual transfer. Søgaard et al. (2019) survey the transfer in the following tasks: document classification, dependency parsing, POS tagging, named entity recognition, super-sense tagging, semantic parsing, discourse parsing, dialogue state tracking, entity linking (wikification), sentiment analysis, machine translation, natural language interference, etc. For example, Ranasinghe and Zampieri (2020) apply large pretrained models in a similar way as we but use offensive language domain and only four languages from different families (English, Spanish, Bengali, and Hindu). In sentiment analysis, which is of particular interest in this work, Mogadala and Rettinger (2016) evaluate their embeddings on the multilingual Amazon product review dataset. In the Twitter sentiment analysis, Wehrmann et al. (2017) use LSTM networks but first learn a joint representation for four languages (English, German, Portuguese, and Spanish) with character-based convolutional neural networks.

## 3. DATASETS AND EXPERIMENTAL SETTINGS

This section presents the evaluation metrics, experimental data, and implementation details of the used neural prediction models.

### 3.1. Evaluation metrics

Following Mozetič et al. (2016), we report the $\bar{F}_1$ score and classification accuracy (*CA*). The $F_1(c)$ score for class value $c$ is the harmonic mean of precision $p$ and recall $r$ for the given class $c$, where the precision is defined as the proportion of correctly classified instances from the instances predicted to be from the class $c$, and the recall is the proportion of correctly classified instances actually from the class $c$:

$$F_1(c) = \frac{2 p_c r_c}{p_c + r_c}.$$

The $F_1$ score returns values from the [0,1] interval, where 1 means perfect classification, and 0 indicates that either precision or recall for class $c$ is 0. We use an instance of the $F_1$ score specifically designed to evaluate the 3-class sentiment models (Kiritchenko et al., 2014). $\bar{F}_1$ is defined as the average over the positive (+) and negative (−) sentiment class:

$$\bar{F}_1 = \frac{F_1(+) + F_1(-)}{2}.$$

$\bar{F}_1$ implicitly considers the ordering of sentiment values by considering only the extreme labels, positive (+) and negative (-). The middle, neutral, is taken into account indirectly. $\bar{F}_1$=1 implies that all negative and positive tweets were correctly classified, and as a consequence, all neutrals as well. $\bar{F}_1$ = 0 indicates that all tweets were classified as neutral, and consequently, all negative and positive tweets were incorrectly classified.

$\bar{F}_1$ is not the best performance measure. First, taking the arithmetic average of the $F_1$ scores over different classes (called macro $F_1$) is methodologically misguided (Flach and Kull, 2015). It is justified only when the class distribution is approximately even, as in our case. Second, $\bar{F}_1$ does not account for correct classifications by chance. A more appropriate measure that allows for class ordering, classification by chance, and class labelling with disagreements is Krippendorff's alpha-reliability (Krippendorff, 2013). However, since $\bar{F}_1$ is commonly used in the sentiment classification community, and the results are typically well-correlated with the alpha-reliability, we decided to report our experimental results in terms of $\bar{F}_1$.

The second score we report is the classification accuracy CA, defined as the ratio of correctly predicted tweets $N_c$ to all the tweets $N$:

$$CA = \frac{N_c}{N}.$$

### 3.2. Datasets

We use a corpus of Twitter sentiment datasets (Mozetič et al., 2016), consisting of 15 languages, with over 1.6 million annotated tweets. The languages covered are Albanian, Bosnian, Bulgarian, Croatian, English, German, Hungarian, Polish, Portuguese, Russian, Serbian, Slovak, Slovene, Spanish, and Swedish. The authors studied the annotators' agreement on the labelled tweets. They discovered that the SVM classifier achieves a significantly lower score for some languages (English, Russian, Slovak) than the annotators. This hints that there might be room for improvement for these languages using a better classification model.

We cleaned the above datasets by removing the duplicated tweets, weblinks, and hashtags. Due to the low quality of sentiment annotations indicated by low self-agreement and low inter-annotator agreement, we removed Albanian and Spanish datasets. For these two languages, the self-agreement expressed with $\bar{F}_1$ score is 0.60 and 0.49, respectively; the inter-annotator agreement is 0.41 and 0.42. As defined above, $\bar{F}_1$ is the arithmetic average of $F_1$ scores for the positive and negative tweets, where $F_1(c)$ is the fraction of equally labelled tweets out of all the tweets with the label $c$.

In the paper where the datasets were introduced (Mozetič et al., 2016), Serbian, Croatian, and Bosnian tweets were merged into a single dataset. The three languages are very similar and difficult to distinguish in short Twitter posts. However, in the study, it turned out that this merge resulted in a poor classification performance due to a very different quality of annotations. In particular, Serbian (71,721 tweets) was annotated by 11 annotators, where two of them accounted for over 40% of the annotations. All the inter-annotator agreement measures come from the Serbian only (1,880 tweets annotated twice by different annotators, $\bar{F}_1$ is 0.51), and there are very few tweets annotated twice by the same annotator (182 tweets only, $\bar{F}_1$ for the self-agreement is 0.46). In contrast, all the Croatian and Bosnian tweets were annotated by a single annotator with high self-agreement estimates. There are 84,001 Croatian tweets, 13,290 annotated twice, and the self-agreement $\bar{F}_1$ is 0.83. There are 38,105 Bosnian tweets, 6,519 annotated twice, and the self-agreement $\bar{F}_1$ is 0.78. The authors concluded that the Croatian and Bosnian tweets' annotation quality is considerably higher than that of the Serbian. When authors constructed separate sentiment classifiers for each language, they observed that individual classifiers are better than the joint Serbian/Croatian/Bosnian model. This paper follows the authors' suggestion that datasets with no overlapping annotations and different annotation quality are better not merged. As a consequence, the Serbian, Croatian, and Bosnian datasets are analysed separately. The characteristics of all the 13 datasets are presented in Table 1.

In general, the quality of the used datasets is comparable to Twitter sentiment datasets used in other studies and sufficient to obtain relevant results in both mono- and cross-lingual comparisons.

|  | Number of tweets | | | | Agreement ($\bar{F}_1$) | |
| --- | --- | --- | --- | --- | --- | --- |
| Language | Negative | Neutral | Positive | All | Self- | Inter- |
| Bosnian | 12,868 | 11,526 | 13,711 | 38,105 | 0.78 | - |
| Bulgarian | 15,140 | 31,214 | 20,815 | 67,169 | 0.77 | 0.50 |
| Croatian | 21,068 | 19,039 | 43,894 | 84,001 | 0.83 | - |
| English | 26,674 | 46,972 | 29,388 | 103,034 | 0.79 | 0.67 |
| German | 20,617 | 60,061 | 28,452 | 109,130 | 0.73 | 0.42 |
| Hungarian | 10,770 | 22,359 | 35,376 | 68,505 | 0.76 | - |
| Polish | 67,083 | 60,486 | 96,005 | 223,574 | 0.84 | 0.67 |
| Portuguese | 58,592 | 53,820 | 44,981 | 157,393 | 0.74 | - |
| Russian | 34,252 | 44,044 | 29,477 | 107,773 | 0.82 | - |
| Serbian | 24,860 | 30,700 | 16,161 | 71,721 | 0.46 | 0.51 |
| Slovak | 18,716 | 14,917 | 36,792 | 70,425 | 0.77 | - |
| Slovene | 38,975 | 60,679 | 34,281 | 133,935 | 0.73 | 0.54 |
| Swedish | 25,319 | 17,857 | 15,371 | 58,547 | 0.76 | - |

Table 1: The left-hand side reports the number of tweets from each category and the overall number of instances for individual languages. The right-hand side contains self-agreement of annotators and inter-annotator agreement for tried languages where more than one annotator was involved.

### 3.3. Implementation details

In our experiments, we use three different types of prediction models, BiLSTM neural networks using joint vector space embeddings constructed with the LASER library, and two variants of BERT, mBERT, and CSE BERT. The original mBERT (bert-multi-cased) is pretrained on 104 languages, has 12 transformer layers and 110 million parameters. The CSE BERT uses the same architecture but is pretrained only on Croatian, Slovene, and English. In the construction of sentiment classification models, we fine-tune the whole network, using the batch size of 32, 2 epochs, and Adam optimiser (Kingma and Ba, 2015). We also tested larger numbers of epochs and larger batch sizes in preliminary experiments, but this did not improve the performance.

The cross-lingual embeddings from the LASER library are pretrained on 93 languages, using BiLSTM networks, and are stored as 1024 dimensional embedding vectors. Our classification models contain an embedding layer, followed by a multilayer perceptron hidden layer of size 8, and an output layer with three neurons (corresponding to three output classes, negative, neutral, and positive sentiment) using the softmax. We use the ReLU activation function and Adam optimiser. The fine-tuning uses a batch size of 32 and 10 epochs.

Further technical details are available in the freely available source code.

### 4. EXPERIMENTS AND RESULTS

Our experimental work focuses on model transfer with cross-lingual embeddings. However, to first establish the performance of different embedding spaces for Twitter sentiment classification, we start with their comparison in a monolingual setting in Section 4.1. We compare the three neural approaches presented in Section 3.3 (common vector space of LASER, mBERT, and CSE BERT). As a baseline, we use the classical approach using bag-of-ngram representation with the SVM classifier. In the cross-lingual experiments, we focus on the two most-successful types of model transfer, described in Sections 2.2 and 2.3: the common vector space of the LASER library and the variants of the multilingual BERT model (mBERT and CSE BERT). We conducted several cross-lingual transfer experiments: transfer of models between languages from the same (Section 4.2) and different language families (Section 4.3), as well as the expansion of training sets with varying amounts of data from other languages (Section 4.4). We end the section with the demonstration of error analysis difficulties (Section 4.5). In the experiments, we did not systematically test all possible combinations of languages and language groups as this would require an excessive amount of computational time and reporting space and would not contribute to the paper's clarity. Instead, we arbitrarily selected a representative set of language combinations in advance. We leave a comprehensive systematic approach based on informative features (Lin et al., 2019) for further work.

### 4.1. Comparing embedding spaces

To establish the performance of different embedding approaches for our Twitter sentiment classification task, we start with experiments in a monolingual setting. We compare the joint vector space embeddings (obtained with the LASER library) with mBERT and CSE BERT. Note that there is no transfer between different languages in this experiment but only a representation (i.e. embeddings) test. To make the results comparable with previous work on these datasets, we report results obtained with 10-fold, stratified, blocked cross-validation. There is no randomisation of training examples in the blocked cross-validation, and each fold is a block of consecutive tweets. Using SVM models, Mozetič et al. (2018) observed that standard cross-validation with a random selection of examples yields too optimistic estimates of classifier performance and should not be used to evaluate classifiers in time-ordered data scenarios. In our study, we used neural networks and conducted a few preliminary experiments with blocked and non-blocked cross-validation but did not observe relevant differences in the obtained scores.

As a baseline, we report the results of majority classifiers and SVM models without neural embeddings (Mozetič et al., 2016). The SVM preprocessing is slightly different from the preprocessing for BERT models (described in Section 3.2) and involves tokenization, stemming/lemmatization, unigram and bigram construction, and elimination of terms that do not appear at least five times in a dataset. A Twitter-specific preprocessing is applied, i.e. replacing URLs, Twitter usernames and hashtags with common tokens, adding emoticon features for different types of emoticons in tweets, handling repetitive letters, etc. The feature vectors are finally constructed by the Delta TF-IDF weighting scheme. As the datasets for the Bosnian, Croatian, and Serbian languages were merged in (Mozetič et al., 2016) due to the similarity of these languages, we report the performance on the merged dataset for the SVM classifier. Results are presented in Table 2.

The SVM baseline using bag-of-ngrams (i.e. bag-of-tokens) representation mostly achieves slightly lower predictive performance than the two neural embedding approaches. We speculate that the main reason is more information about the language structure contained in precomputed dense embeddings used by the neural approaches. Together with the fact that standard feature-based machine learning approaches require much more preprocessing effort, it seems that there are no good reasons why to bother with this approach in text classification; we, therefore, omit this method from further experiments. The mBERT model is the best of the tested methods, achieving the best $\overline{F_1}$ and CA scores in six languages (in bold), closely followed by the LASER approach, which achieves the best $\overline{F_1}$ score in five languages and the best CA score in three languages. The CSE BERT is specialised in only three languages, and it achieves the best scores in languages where it is trained (except in English, where it is close behind mBERT) and in Bosnian, which is similar to Croatian.

|  | LASER |  | mBERT |  | CSE BERT |  | SVM |  | Majority |
|---|---|---|---|---|---|---|---|---|---|
| Language | $\bar{F}_1$ | CA | $\bar{F}_1$ | CA | $\bar{F}_1$ | CA | $\bar{F}_1$ | CA | CA |
| Bosnian | **0.68** | 0.64 | 0.65 | 0.60 | **0.68** | **0.65** | (0.61 | 0.56) | 0.36 |
| Bulgarian | 0.53 | **0.59** | **0.58** | **0.59** | 0.00 | 0.45 | 0.52 | 0.54 | 0.46 |
| Croatian | 0.72 | 0.68 | 0.64 | 0.66 | **0.76** | **0.71** | (0.61 | 0.56) | 0.52 |
| English | 0.62 | 0.65 | **0.68** | **0.68** | 0.67 | 0.66 | 0.63 | 0.64 | 0.44 |
| German | 0.52 | 0.64 | **0.66** | **0.66** | 0.31 | 0.59 | 0.54 | 0.61 | 0.53 |
| Hungarian | 0.63 | 0.67 | **0.65** | **0.69** | 0.57 | 0.65 | 0.64 | 0.67 | 0.53 |
| Polish | **0.70** | 0.66 | **0.70** | **0.70** | 0.56 | 0.57 | 0.68 | 0.63 | 0.44 |
| Portuguese | 0.48 | 0.47 | 0.50 | 0.49 | 0.12 | 0.22 | **0.55** | **0.51** | 0.37 |
| Russian | **0.70** | **0.70** | 0.64 | 0.64 | 0.07 | 0.43 | 0.61 | 0.60 | 0.40 |
| Serbian | 0.50 | 0.54 | 0.50 | 0.52 | 0.30 | 0.50 | (**0.61** | **0.56**) | 0.43 |
| Slovak | **0.72** | **0.72** | 0.67 | 0.66 | 0.69 | 0.71 | 0.68 | 0.68 | 0.52 |
| Slovene | 0.57 | 0.58 | 0.58 | 0.58 | **0.60** | **0.61** | 0.55 | 0.54 | 0.43 |
| Swedish | **0.67** | 0.64 | **0.67** | **0.65** | 0.54 | 0.56 | 0.66 | 0.62 | 0.43 |
| #Best | 5 | 3 | 6 | 6 | 3 | 3 | 2 | 2 | 0 |
| Average | 0.62 | 0.63 | 0.62 | 0.62 | 0.45 | 0.56 | 0.61 | 0.60 | 0.45 |

Table 2: Comparison of different representations: supervised mapping into a joint vector space with the LASER library, mBERT, CSE BERT, and bag-of-ngrams with the SVM classifier. The best score for each language and metric is in bold. As a baseline, we give the classification accuracy of the majority classifier ($\bar{F}_1$ is zero or undefined for this classifier). In the last two rows, we present the count of best scores and average scores for each model. The SVM results for Bosnian, Croatian, and Serbian were obtained with the model trained on the merged dataset of these languages model and are therefore not directly compatible with the language-specific results for the other representations.

Overall, it seems that large pretrained transformer models (mBERT and CSE BERT) are dominating in the Twitter sentiment prediction. The downside of these models is that their training, fine-tuning, and execution require more computational time than precomputed fixed embeddings. Nevertheless, with progress in optimisation techniques for neural network learning and the advent of computationally more efficient BERT variants, e.g., (You et al., 2020), this obstacle might disappear in the future.

4.2. **Transfer to the same language family**

The transfer of prediction models between similar languages from the same language family is the most likely to be successful. We test several combinations of source and target languages from Slavic and Germanic language families. We report the results in Table 3.

In each experiment, we use the entire dataset(s) of the source language as the training set and the whole dataset of the target language as the testing set, i.e. we do a zero-shot transfer. We compare the results with the LASER embeddings with the BiLSTM network using training and testing set from the target language, where 70% of the dataset is used for training and 30% for testing. As we use large datasets, the latter results can be taken as an upper bound of what cross-lingual transfer models could achieve in ideal conditions. The baseline CA for each of the target languages is contained in Table 2.

Table 3 (bottom line) shows that there is a gap in the performance of transfer learning models and native models. On average, the gap in $\overline{F}_1$ is 5% for the LASER approach, 6% for mBERT, and 8% for CSE BERT. For CA, the average gap is 7% for both LASER and mBERT and 8% for CSE BERT. As seen in Table 2, the CA of the majority classifier is on average 17% lower than LASER and mBERT. There are significant differences between languages, and we advise testing both LASER and mBERT for a specific new language, as the models are highly competitive. The CSE BERT is slightly less successful measured with the average performance gap over all languages as the gap is 8% in both $\overline{F}_1$ and CA. However, if we take only the three languages used in CSE BERT (Croatian, Slovene, and English) as shown in Table 4, conclusions are entirely different. The average performance gap is 0% in $\overline{F}_1$ and 1% in the classification accuracy, meaning that we get almost a perfect cross-lingual transfer for these languages on the Twitter sentiment prediction task.

We also tried more than one input language at once, for example, German and Swedish as source languages and English as the target language, as shown in Table 3. The success of the tested combinations is mixed: for some models and some languages, we slightly improve the scores, while for others, we slightly decrease them. We hypothesise that our datasets for individual languages are large enough so that adding additional training data does not help.

|  |  | LASER | | mBERT | | CSE BERT | | Both target | |
|---|---|---|---|---|---|---|---|---|---|
| Source | Target | $\overline{F}_1$ | CA | $\overline{F}_1$ | CA | $\overline{F}_1$ | CA | $\overline{F}_1$ | CA |
| German | English | 0.55 | 0.59 | **0.63** | **0.64** | 0.42 | 0.42 | 0.62 | 0.65 |
| English | German | 0.55 | 0.60 | **0.66** | **0.70** | 0.50 | 0.58 | 0.53 | 0.65 |
| Polish | Russian | **0.64** | **0.59** | 0.57 | 0.57 | 0.50 | 0.40 | 0.70 | 0.70 |
| Polish | Slovak | **0.63** | 0.59 | 0.58 | 0.59 | **0.63** | **0.65** | 0.72 | 0.72 |
| German | Swedish | 0.58 | 0.57 | **0.59** | **0.59** | 0.58 | 0.56 | 0.67 | 0.65 |
| German Swedish | English | **0.58** | **0.60** | 0.55 | 0.56 | 0.41 | 0.42 | 0.62 | 0.65 |
| Slovene Serbian | Russian | 0.53 | 0.55 | 0.57 | **0.57** | **0.58** | 0.48 | 0.70 | 0.70 |
| Slovene Serbian | Slovak | **0.59** | 0.52 | 0.57 | 0.59 | 0.48 | **0.60** | 0.72 | 0.72 |
| Serbian | Slovene | 0.54 | **0.57** | 0.54 | 0.54 | **0.56** | 0.55 | 0.60 | 0.60 |
| Serbian | Croatian | **0.67** | 0.64 | 0.65 | 0.62 | 0.65 | **0.70** | 0.73 | 0.68 |
| Serbian | Bosnian | **0.65** | 0.61 | 0.61 | 0.60 | 0.59 | **0.62** | 0.67 | 0.64 |
| Polish | Slovene | 0.51 | 0.48 | **0.55** | **0.54** | 0.50 | 0.53 | 0.60 | 0.60 |
| Slovak | Slovene | 0.52 | 0.51 | 0.54 | 0.54 | **0.58** | **0.58** | 0.60 | 0.60 |
| Croatian | Slovene | 0.53 | 0.53 | 0.53 | 0.54 | **0.61** | **0.60** | 0.60 | 0.60 |
| Croatian | Serbian | **0.54** | **0.52** | 0.52 | 0.51 | 0.52 | 0.49 | 0.48 | 0.54 |
| Croatian | Bosnian | 0.66 | 0.61 | 0.57 | 0.56 | **0.67** | **0.62** | 0.67 | 0.64 |
| Slovene | Croatian | 0.70 | 0.65 | 0.64 | 0.63 | **0.73** | **0.69** | 0.73 | 0.68 |
| Slovene | Serbian | **0.52** | **0.55** | 0.46 | 0.49 | 0.47 | 0.50 | 0.48 | 0.54 |
| Slovene | Bosnian | **0.66** | 0.61 | 0.58 | 0.56 | 0.66 | **0.62** | 0.67 | 0.64 |
| Average performance gap | | 0.05 | 0.07 | 0.06 | 0.07 | 0.08 | 0.08 | | |

Table 3: The transfer of trained models between languages from the same language family using LASER common vector space, mBERT, and CSE BERT. We compare the results with both training and testing set from the target language using the LASER approach (the right-most two columns).

|  |  | LASER | | mBERT | | CSE BERT | | Both target | |
| --- | --- | --- | --- | --- | --- | --- | --- | --- | --- |
| Source | Target | $\bar{F}_1$ | CA | $\bar{F}_1$ | CA | $\bar{F}_1$ | CA | $\bar{F}_1$ | CA |
| Croatian | Slovene | 0.53 | 0.53 | 0.53 | 0.54 | **0.61** | **0.60** | 0.60 | 0.60 |
| Croatian | English | **0.63** | 0.63 | **0.63** | **0.66** | 0.62 | 0.64 | 0.62 | 0.65 |
| English | Slovene | 0.54 | 0.57 | 0.50 | 0.53 | **0.59** | **0.57** | 0.60 | 0.60 |
| English | Croatian | 0.62 | **0.67** | 0.67 | 0.63 | **0.73** | **0.67** | 0.73 | 0.68 |
| Slovene | English | 0.63 | 0.64 | **0.65** | **0.67** | 0.63 | 0.64 | 0.62 | 0.65 |
| Slovene | Croatian | 0.70 | 0.65 | 0.64 | 0.63 | **0.73** | **0.69** | 0.73 | 0.68 |
| Croatian English | Slovene | 0.54 | 0.54 | 0.53 | 0.54 | **0.60** | **0.58** | 0.60 | 0.60 |
| Croatian Slovene | English | 0.62 | 0.61 | **0.65** | **0.67** | 0.63 | 0.65 | 0.62 | 0.65 |
| English Slovene | Croatian | 0.64 | 0.68 | 0.63 | 0.63 | **0.68** | **0.70** | 0.73 | 0.68 |
| Average performance gap | | 0.04 | 0.03 | 0.04 | 0.03 | 0.00 | 0.01 | | |

Table 4: The transfer of sentiment models between all combinations of languages on which CSE BERT was trained (Croatian, Slovene, and English).

Another baseline approach we did not test in our study is the translation of the training or testing set instead of using cross-lingual representations. Conneau et al. (2018) showed that this could be a very competitive approach, mainly when translating testing sets, but strongly depends on translations' quality. Translation-based approaches are computationally much more intensive in the case of machine translation or slow and expensive in the case of human translation. The quality of machine translation approaches depends on the translation error, changes in style, and introduced artifacts that result in discrepancies between the training and testing data.

### 4.3. Transfer to a different language family

The transfer of prediction models between languages from different language families is less likely to be successful. Nevertheless, to observe the difference, we test several combinations of source and target languages from different language families (one from Slavic, the other from Germanic, and vice-versa). We compare the LASER approach with mBERT models; the CSE BERT is not constructed for this setting, and we skip it in this experiment. We report the results in Table 5.
The results show that with the LASER approach, there is an average decrease of performance for transfer learning models of 11% (both $\bar{F}_1$ and CA), and for mBERT, the gap is 9%. This gap is significant and makes the resulting transferred models less useful in the target languages, though there are considerable differences between the languages. As seen in Table 2, the CA of the majority classifier is 17% less than LASER and mBERT.

### 4.4. Increasing datasets with several languages

Another type of cross-lingual transfer is possible if we increase the training sets with instances from several related and unrelated languages. We conduct two sets of experiments in this scenario. In the first setting, reported in Table 6, we constructed the training set in each experiment with instances from several languages and 70% of the target language dataset. The remaining 30% of target language instances are used as the testing set. In the second setting, reported in Table 7, we merge *all* other languages and 70% of the target language into a joint training set. We compare the LASER approach, mBERT, and also CSE BERT, as Slovene and Croatian are involved in some combinations.

|  |  | LASER | | mBERT | | Both target | |
| --- | --- | --- | --- | --- | --- | --- | --- |
| Source | Target | $\bar{F}_1$ | CA | $\bar{F}_1$ | CA | $\bar{F}_1$ | CA |
| Russian | English | **0.52** | 0.56 | **0.52** | **0.57** | 0.62 | 0.65 |
| English | Russian | **0.57** | **0.58** | 0.55 | 0.57 | 0.70 | 0.70 |
| English | Slovak | 0.46 | 0.44 | **0.57** | **0.58** | 0.72 | 0.72 |
| Polish, Slovene | English | 0.58 | 0.57 | **0.60** | **0.60** | 0.62 | 0.65 |
| German, Swedish | Russian | 0.61 | **0.61** | **0.62** | 0.59 | 0.70 | 0.70 |
| English, German | Slovak | 0.50 | 0.47 | **0.56** | **0.54** | 0.72 | 0.72 |
| German | Slovene | **0.54** | **0.56** | 0.53 | 0.54 | 0.60 | 0.60 |
| English | Slovene | **0.54** | **0.57** | 0.50 | 0.53 | 0.60 | 0.60 |
| Swedish | Slovene | **0.54** | **0.56** | 0.52 | 0.54 | 0.60 | 0.60 |
| Hungarian | Slovene | 0.52 | 0.52 | **0.53** | **0.54** | 0.60 | 0.60 |
| Portuguese | Slovene | 0.51 | 0.49 | **0.54** | **0.54** | 0.60 | 0.60 |
| Average performance gap | | 0.11 | 0.11 | 0.09 | 0.09 | | |

Table 5: The transfer of trained models between languages from different language families using LASER common vector space and mBERT. We compare the results with both training and testing set from the target language using the LASER approach (the right-most two columns).

Table 6 shows a gap between learning models using the expanded datasets and models with only target language data. The decrease is more extensive for both BERT models (on average around 10%) than for the LASER approach (the decrease is on average 3% for $\bar{F}_1$ and 5% for CA). These results indicate that the tested expansion of datasets was unsuccessful, i.e. the provided amount of training instances in the target language was already sufficient for successful learning. We hypothesize that the additional instances from other languages in the transformed space are likely to be of a different style or use different topics than the native instances and therefore decrease the performance. However, this issue warrants further investigation.

|  |  | LASER | | mBERT | | CSEBERT | | Target only | |
| --- | --- | --- | --- | --- | --- | --- | --- | --- | --- |
| Source | Target | $\bar{F}_1$ | CA | $\bar{F}_1$ | CA | $\bar{F}_1$ | CA | $\bar{F}_1$ | CA |
| English, Croatian, Slovene | Slovene | 0.58 | 0.53 | 0.46 | 0.45 | **0.60** | **0.58** | 0.60 | 0.60 |
| English, Croatian, Serbian, Slovak | Slovak | **0.67** | **0.65** | 0.57 | 0.54 | 0.27 | 0.37 | 0.72 | 0.72 |
| Hungarian, Slovak, English, Croatian, Russian | Russian | **0.67** | **0.65** | 0.61 | 0.59 | 0.63 | 0.61 | 0.70 | 0.70 |
| Russian, Swedish, English | English | 0.60 | 0.61 | **0.62** | 0.60 | 0.59 | **0.62** | 0.62 | 0.65 |
| Croatian, Serbian, Bosnian, Slovene | Slovene | 0.54 | **0.58** | 0.44 | 0.45 | **0.57** | 0.56 | 0.60 | 0.60 |
| English, Swedish, German | German | 0.55 | 0.60 | **0.60** | **0.64** | 0.47 | 0.58 | 0.53 | 0.65 |
| Average performance gap | | 0.03 | 0.05 | 0.08 | 0.11 | 0.11 | 0.10 | | |

Table 6: The expansion of training sets with instances from several languages. We compare the LASER approach, mBERT, and CSE BERT. As the upper bound, we give results of the LASER approach trained on only the target language.

The results in Table 7, where we test the expansion of the training set (consisting of 70% of the dataset in the target language) with all other languages, show that using many languages and significant enlargement of datasets is also not successful. The two improvements in the LASER approach over using only target language are limited to a single metric ($F_1$ in the case of Bulgarian and Serbian) which indicates that true positives are favoured at the expense of true negatives. For all the other languages, the tried expansions of training sets are unsuccessful for the LASER approach; the difference to native models is on average 3.5% for the $\overline{F}_1$ score and 6% for the CA. The mBERT models are in almost all cases more successful in this massive transfer than LASER models, and they sometimes marginally beat the reference mBERT approach trained only on the target language.

| Target | LASER All & Target $\overline{F}_1$ | LASER All & Target CA | LASER Only Target $\overline{F}_1$ | LASER Only Target CA | mBERT All & Target $\overline{F}_1$ | mBERT All & Target CA | mBERT Only Target $\overline{F}_1$ | mBERT Only Target CA |
|---|---|---|---|---|---|---|---|---|
| Bosnian | 0.64 | 0.59 | 0.67 | 0.64 | 0.63 | 0.60 | 0.65 | 0.60 |
| Bulgarian | **0.54** | 0.56 | 0.50 | 0.59 | **0.60** | **0.60** | 0.58 | 0.59 |
| Croatian | 0.63 | 0.57 | 0.73 | 0.68 | **0.65** | 0.63 | 0.64 | 0.66 |
| English | 0.58 | 0.60 | 0.62 | 0.65 | 0.64 | **0.69** | 0.68 | 0.68 |
| German | 0.52 | 0.59 | 0.53 | 0.65 | 0.61 | 0.66 | 0.66 | 0.66 |
| Hungarian | 0.59 | 0.61 | 0.60 | 0.67 | 0.65 | 0.69 | 0.65 | 0.69 |
| Polish | 0.67 | 0.63 | 0.70 | 0.66 | **0.71** | **0.71** | 0.70 | 0.70 |
| Portuguese | 0.44 | 0.39 | 0.52 | 0.51 | **0.52** | **0.52** | 0.50 | 0.49 |
| Russian | 0.66 | 0.64 | 0.70 | 0.70 | **0.67** | **0.66** | 0.64 | 0.64 |
| Serbian | **0.52** | 0.49 | 0.48 | 0.54 | **0.53** | 0.51 | 0.50 | 0.52 |
| Slovak | 0.64 | 0.61 | 0.72 | 0.72 | 0.67 | 0.65 | 0.67 | 0.66 |
| Slovene | 0.54 | 0.50 | 0.60 | 0.60 | 0.56 | 0.54 | 0.58 | 0.58 |
| Swedish | 0.63 | 0.59 | 0.67 | 0.65 | 0.67 | 0.64 | 0.67 | 0.65 |
| Avg. gap | 0.03 | 0.06 | | | 0.00 | 0.00 | | |

Table 7: The expansion of training sets with instances from all other languages (+70% of the target language instances) to train the LASER approach and mBERT. We compare the results with the training on only the target language. The scores where models with the expanded training sets beat their respective reference scores are in bold.

### 4.5 Error analysis attempt

A potential limitation of our analyses is that they are purely statistical as we did not do any manual error analysis. However, for the type and amount of data analysed in our study, manual error analysis is likely impossible. Relatively low self- and inter-annotator agreements, reported in Table 1, indicate that Twitter sentiment annotation is difficult due to short messages and insufficient context. However, our statistical results show that the most successful models are on the level of inter-annotator agreement. For most of sentiment analysis applications this is sufficient.

To illustrate the difficulty and avoid cherry-picking, we provide the first 40 randomly selected tweets from our English dataset in Table 8. We supply the (subjective) ground truth and predictions of two mBERT models. The first mBERT one was trained on the English dataset ($\overline{F}_1 = 0.68, CA = 0.68$, see Table 2) and the second one was trained on Slovene ($\overline{F}_1 = 0.65, CA = 0.67$, see Table 4). Unfortunately, we cannot extract any meaningful pattern why the predictions differ among themselves or from the ground truth. The statistical analysis of the performance is likely to be the only sensible approach.

| #  | T | E | S | Tweets |
|----|---|---|---|--------|
| 1  | + | + | - | At the Krishna temple in east Dallas. These folks party hard |
| 2  | - | = | = | I can hold on to something for so long |
| 3  | = | + | + | Do I fit in the pic ? Cx |
| 4  | - | - | = | Playing Madden Till I Take My Boy To Work ! #3rdShiftNiggas ! |
| 5  | + | = | - | PURRRRFECTION |
| 6  | + | + | - | They talk so freaking loud here.. |
| 7  | = | = | + | We spend our lives searching for answers we don't know |
| 8  | + | + | = | GOOOOOOOOOOOOOOOLLL Lllll |
| 9  | = | = | = | 😂😂 I'm padaziling on mfs 💫💫 |
| 10 | - | = | + | I miss bae she needa text me |
| 11 | = | + | + | yeah |
| 12 | = | = | = | NEW YORK ON 🔥🔥🔥 ANOTHER HOT NIGGA OUTTA NEW YORK 🔥🔥🔥💯💯💣💣💣 |
| 13 | + | + | - | I notice everything but say nothing |
| 14 | = | = | - | 3 months until 2015 it's ridiculous how quick this year has gone |
| 15 | - | - | + | It was cool seeing William after not seeing him for a while 😊💕 ft. Mike 😂 |
| 16 | = | - | - | I guess. Rich brown eyes can send you into a trance really. Love brown eyes. |
| 17 | + | + | = | yaku is so Torn btwn telling lev to 'go home and change' and thinking Fuck it its Lev this is what i signed up for |
| 18 | - | - | - | all i see is booty  all i see is booty |
| 19 | = | - | + | i'm sorry 😣😂 |
| 20 | - | = | - | FAIR HELP EVERY DAY THIS WEEK AFTER SCHOOL ! COME OUT AND HELP US WIN FAIR ! |
| 21 | + | - | = | Always remember |
| 22 | + | + | = | Have you ever wanted to be a princess? Just leave your boyfriend and marry me. Today! |
| 23 | = | = | + | I haven't study this whole weekend 😩 |
| 24 | - | + | - | watching WWE Night of Champions 2014 |
| 25 | + | + | = | Woah your avi is just too 💣😍🔥👌 & is that a leg I see there |
| 26 | - | - | = | him but had to proceed to open Ambrose's return |
| 27 | = | = | = | I can't wait for the state fair😍📈    I can't wait for the state fair😍📈 |
| 28 | = | = | + | how much wood could a woodchuck chuck if you sent me nudes ;-) — Goodbye. " ;-( |
| 29 | + | + | + | Im just waiting for u to respond |
| 30 | + | - | + | I scored a goal today⚽👌 |
| 31 | = | = | = | i have a test tomorrow and i studied for about a good 30 seconds i should be good |
| 32 | - | + | = | Well you aren't texting me back so I made you notice me 😤😏 |
| 33 | - | = | + | You determine how far you go in life |
| 34 | - | = | + | Orange Is the New Agos. #TheGoodWife |
| 35 | - | - | = | I love you too |
| 36 | + | = | = | Nobody is ever on yahoo messenger |
| 37 | - | = | - | Determining extent of corn frost damage: Inner shank tissue should be cream color/firm for sugar movement to continue ht… |
| 38 | = | + | = | I can't stand 95% of people at McCracken. There's no need to talk crap to every school we play in every sport. It makes all… |
| 39 | + | - | + | 🙈👌💯 'That's 9 cent worth of chicken you just threw away' 😩 |
| 40 | = | = | + | Are you a fan of Jackson? RT Fan Fav Nah |

Table 8: The first 40 randomly selected tweets from the English dataset. The column labels represent the ground truth (T), prediction with mBERT trained on 70% of English dataset (E), prediction with mBERT trained on full Slovene and 70% of English dataset (S), and the text of tweets (Tweets). The labels +, -, and = stand for positive, negative, and neutral sentiment.

## 5. CONCLUSIONS

We studied state-of-the-art approaches to the cross-lingual transfer of Twitter sentiment prediction models: mappings of words into the common vector space using the LASER library and two multilingual BERT variants (mBERT and trilingual CSE BERT). Our empirical evaluation is based on relatively large datasets of labelled tweets from 13 European languages. We first tested the success of these text representations in a monolingual setting. The results show that BERT variants are the most successful, closely followed by the LASER approach, while the classical bag-of-ngrams representation coupled with the SVM classifier is less competitive with neural approaches. In the cross-lingual experiments, the results show a transfer potential using the models trained on similar languages. Compared to training and testing on the same language, with LASER, we get on average 5% lower $\bar{F}_1$ score and with mBERT 6% lower $\bar{F}_1$ score. The transfer of models with CSE BERT is very successful in the three languages covered by this model, where we get no performance gap compared to the LASER approach trained and tested on the target language. Using models trained on languages from different language families produces larger differences (on average around 10% for $\bar{F}_1$ and CA). Our attempt to expand training sets with instances from different languages was unsuccessful using either additional instances from a small group of languages or instances from all other languages. The source code of our analyses is freely available[3].

In further work, we plan to test different amounts of available target data and introduce the translation baseline. We plan to expand BERT models with additional emotional and subjectivity information. Given the favourable results in cross-lingual transfer, we will expand the work to other relevant tasks.

## ACKNOWLEDGEMENTS

The research was supported by the Slovene Research Agency through research core funding no. P6-0411 and P2-103, as well as project no. J6-2581 (Computer-assisted multilingual news discourse analysis with contextual embeddings). This paper is supported by European Union's Horizon 2020 Programme project EMBEDDIA (Cross-Lingual Embeddings for Less-Represented Languages in European News Media, grant no. 825153), and Rights, Equality and Citizenship Programme project IMSyPP (Innovative Monitoring Systems and Prevention Policies of Online Hate Speech, grant no. 875263). The results of this publication reflect only the authors' view, and the Commission is not responsible for any use that may be made of the information it contains.

## REFERENCES

Artetxe, M., and Schwenk, H. (2019): Massively multilingual sentence embeddings for zero-shot cross-lingual transfer and beyond. *Transactions of the Association for Computational Linguistics*, 7:597–610.

Artetxe, M., Labaka, G., and Agirre, E. (2018a): Generalising and improving bilingual word embedding mappings with a multi-step framework of linear transformations. In *Thirty-Second AAAI Conference on Artificial Intelligence*.

Artetxe, M., Labaka, G., and Agirre, E. (2018b): A robust self-learning method for fully unsupervised cross-lingual mappings of word embeddings. In *Proceedings of the 56th Annual Meeting of the Association for Computational Linguistics (Volume 1: Long Papers)*, pages 789–798.

Bojanowski, P., Grave, E., Joulin, A., and Mikolov, T. (2017): Enriching word vectors with subword information. *Transactions of the Association for Computational Linguistics*, 5:135–146.

---

[3] https://github.com/kristjanreba/cross-lingual-classificationof-tweet-sentiment

Conneau, A., Lample, G., Ranzato, M.A., Denoyer, L., and J'egou, H. (2018): Word translation without parallel data. In *Proceedings of International Conference on Learning Representation (ICLR)*.

Conneau, A., Rinott, R., Lample, G., Williams, A., Bowman, S., Schwenk, H., & Stoyanov, V. (2018). XNLI: Evaluating Cross-lingual Sentence Representations. In *Proceedings of the 2018 Conference on Empirical Methods in Natural Language Processing* (pp. 2475-2485).

Devlin, J., Chang, M.-W., Lee, K., and Toutanova, K. (2019): BERT: Pre-training of deep bidirectional transformers for language understanding. In *Proceedings of the 2019 Conference of the North American Chapter of the Association for Computational Linguistics: Human Language Technologies, Volume 1 (Long and Short Papers)*, pages 4171–4186.

Flach P., and Kull M. (2015): Precision-recall-gain curves: PR analysis done right. In *Advances in Neural Information Processing Systems (NIPS)*, pp. 838-846.

Jianqiang, Z., Xiaolin, G., and Xuejun, Z. (2018): Deep convolution neural networks for Twitter sentiment analysis. *IEEE Access*, 6, 23253-23260.

Kingma, D. P., & Ba, J. (2015). Adam: A method for stochastic optimization. In Proceedings of International Conference for Learning Representations, ICLR 2015.

Kiritchenko, S., Zhu, X., Mohammad, S.M. (2014): Sentiment analysis of short informal texts. *Journal of Artificial Intelligence Research*, 50:723–762.

Krippendorff, K. (2013): *Content Analysis, An Introduction to Its Methodology. 3rd ed*. Thousand Oaks, CA, USA: Sage Publications.

Lin,Y.H., Chen, C.Y., Lee, J., Li, Z., Zhang, Y., Xia, M., Rijhwani, S., He, J., Zhang, Z., Ma, X., et al. (2019): Choosing transfer languages for cross-lingual learning. In *Proceedings of the 57th Annual Meeting of the Association for Computational Linguistics (ACL)*, pages 3125–3135.

Mikolov, T., Le, Q.V., and Sutskever, I (2013): Exploiting similarities among languages for machine translation. *arXiv preprint* 1309.4168.

Mogadala, A., and Rettinger, A. (2016). Bilingual word embeddings from parallel and non-parallel corpora for cross-language text classification. In *Proceedings of NAACL-HLT*, pp. 692–702

Mozetič, I., Grčar, M., and Smailović, J. (2016): Multilingual Twitter sentiment classification: The role of human annotators. *PLOS ONE*, 11(5).

Mozetič, I., Torgo, L., Cerqueira, V., and Smailović, J. (2018) How to evaluate sentiment classifiers for Twitter time-ordered data? *PLoS ONE* 13(3).

Naseem, U., Razzak, I., Musial, K., and Imran, M. (2020). Transformer-based deep intelligent contextual embedding for Twitter sentiment analysis. *Future Generation Computer Systems*, 113, 58-69.

Peters, M., Neumann, M., Iyyer, M., Gardner, M., Clark, C., Lee, K., and Zettlemoyer, L. (2018): Deep contextualised word representations. In *Proceedings of the 2018 Conference of the North American Chapter of the Association for Computational Linguistics: Human Language Technologies, Volume 1 (Long Papers)*, pp. 2227–2237.

Ranasinghe, T., and Zampieri, M. (2020): Multilingual Offensive Language Identification with Cross-lingual Embeddings. In *Proceedings of the 2020 Conference on Empirical Methods in Natural Language Processing (EMNLP)*, pp. 5838-5844.

Rosenthal, S., Nakov, P., Kiritchenko, S., Mohammad, S.M., Ritter, A., and Stoyanov, V. (2015) SemEval-2015 task 10: Sentiment Analysis in Twitter. In *Proceedings of 9th International Workshop on Semantic Evaluation (SemEval)*, pp. 451–463.

Saif, H., Fernández, M., He, Y., Alani, H. (2013): Evaluation datasets for Twitter sentiment analysis: A survey and a new dataset, the STS-Gold. In *1st Intl. Workshop on Emotion and Sentiment in Social and Expressive Media: Approaches and Perspectives from AI (ESSEM)*.

Søgaard, A., Vulić, I., Ruder, S., and Faruqui, M. (2019): *Cross-Lingual Word Embeddings*. Morgan & Claypool Publishers.

Ulčar, M., and Robnik-Šikonja, M. (2020): FinEst BERT and CroSloEngual BERT. In *International Conference on Text, Speech, and Dialogue (TSD)*, pp. 104-111.

Vaswani, A., Shazeer, N., Parmar, N., Uszkoreit, J., Jones, L., Gomez, A.N., Kaiser, Ł., and Polosukhin, I. (2017): Attention is all you need. In *Advances in Neural Information Processing Systems (NIPS)*, pp. 5998–6008.

Virtanen, A., Kanerva, J., Ilo, R., Luoma, J., Luoto-lahti, J., Salakoski, T., Ginter, F., and Pyysalo, S. (2019): Multilingual is not enough: BERT for Finnish. *arXiv preprint* arXiv:1912.07076.

Wehrmann, J., Becker, W., Cagnini, H. E., and Barros, R. C. (2017). A character-based convolutional neural network for language-agnostic Twitter sentiment analysis. In 2017 *International Joint Conference on Neural Networks (IJCNN)*, pp. 2384-2391.

You, Y., Li, J., Reddi, S., Hseu, J., Kumar, S., Bhojanapalli, S., Song, X., Demmel, J., Keutzer, K., and Hsieh, C.J. (2019): Large batch optimization for deep learning: Training BERT in 76 minutes. In *International Conference on Learning Representations (ICLR)*.